\documentclass[10pt,twocolumn,letterpaper]{article}

\usepackage{booktabs}
\usepackage{graphicx}
\usepackage{multirow}
\usepackage{amsmath}

\usepackage{tikz}
\usepackage{geometry}
\usepackage{tikz}
\usetikzlibrary{calc, angles, quotes, arrows.meta, positioning, shapes.geometric, shadows, backgrounds}

\usepackage[pagenumbers]{cvpr} 

\definecolor{cvprblue}{rgb}{0.21,0.49,0.74}
\usepackage[pagebackref,breaklinks,colorlinks,allcolors=cvprblue]{hyperref}

\def\paperID{****} 
\def\confName{CVPR}
\def\confYear{2026}

\title{StereoPilot: Learning Unified and Efficient Stereo Conversion \\ via Generative Priors}

\def\authorBlock{
    Guibao Shen\textsuperscript{$1,3$}\thanks{Equal contribution} \thanks{This work was conducted during the author's internship at Kling.}\qquad
    Yihua Du\textsuperscript{$1$}\footnotemark[1]  \qquad
    Wenhang Ge\textsuperscript{$1,3$}\footnotemark[1] \footnotemark[2]\qquad
    Jing He\textsuperscript{$1$} \quad
    Chirui Chang\textsuperscript{$3$} \quad \\
    Donghao Zhou\textsuperscript{$4$} \quad
    Zhen Yang\textsuperscript{$1$} \quad
    Luozhou Wang\textsuperscript{$1$} \quad
    Xin Tao\textsuperscript{$3$} \quad
    Ying-Cong Chen\textsuperscript{$1,2$} \thanks{Corresponding author}\quad
    
    \\
    \small$^1$ HKUST(GZ)\quad 
    \small$^2$ HKUST\quad 
    \small$^3$ Kling Team, Kuaishou Technology\quad 
    \small$^4$ CUHK \\
    \vspace{0.3cm}
    {\url{https://github.com/KlingTeam/StereoPilot}}
}
\author{\authorBlock}

\begin{document}
\maketitle
\begin{abstract}
The rapid growth of stereoscopic displays, including VR headsets and 3D cinemas, has led to increasing demand for high-quality stereo video content. However, producing 3D videos remains costly and complex, while automatic Monocular-to-Stereo conversion is hindered by the limitations of the multi-stage ``Depth-Warp-Inpaint'' (DWI) pipeline. This paradigm suffers from error propagation, depth ambiguity, and format inconsistency between parallel and converged stereo configurations. To address these challenges, we introduce UniStereo, the first large-scale unified dataset for stereo video conversion, covering both stereo formats to enable fair benchmarking and robust model training. Building upon this dataset, we propose StereoPilot, an efficient feed-forward model that directly synthesizes the target view without relying on explicit depth maps or iterative diffusion sampling. Equipped with a learnable domain switcher and a cycle consistency loss, StereoPilot adapts seamlessly to different stereo formats and achieves improved consistency. Extensive experiments demonstrate that StereoPilot significantly outperforms state-of-the-art methods in both visual fidelity and computational efficiency. Project page:
\href{https://hit-perfect.github.io/StereoPilot/}{StereoPilot Page}.
\end{abstract}    
\section{Introduction}
\label{sec:intro}

\begin{figure}
    \centering
    \includegraphics[width=1.0\linewidth]{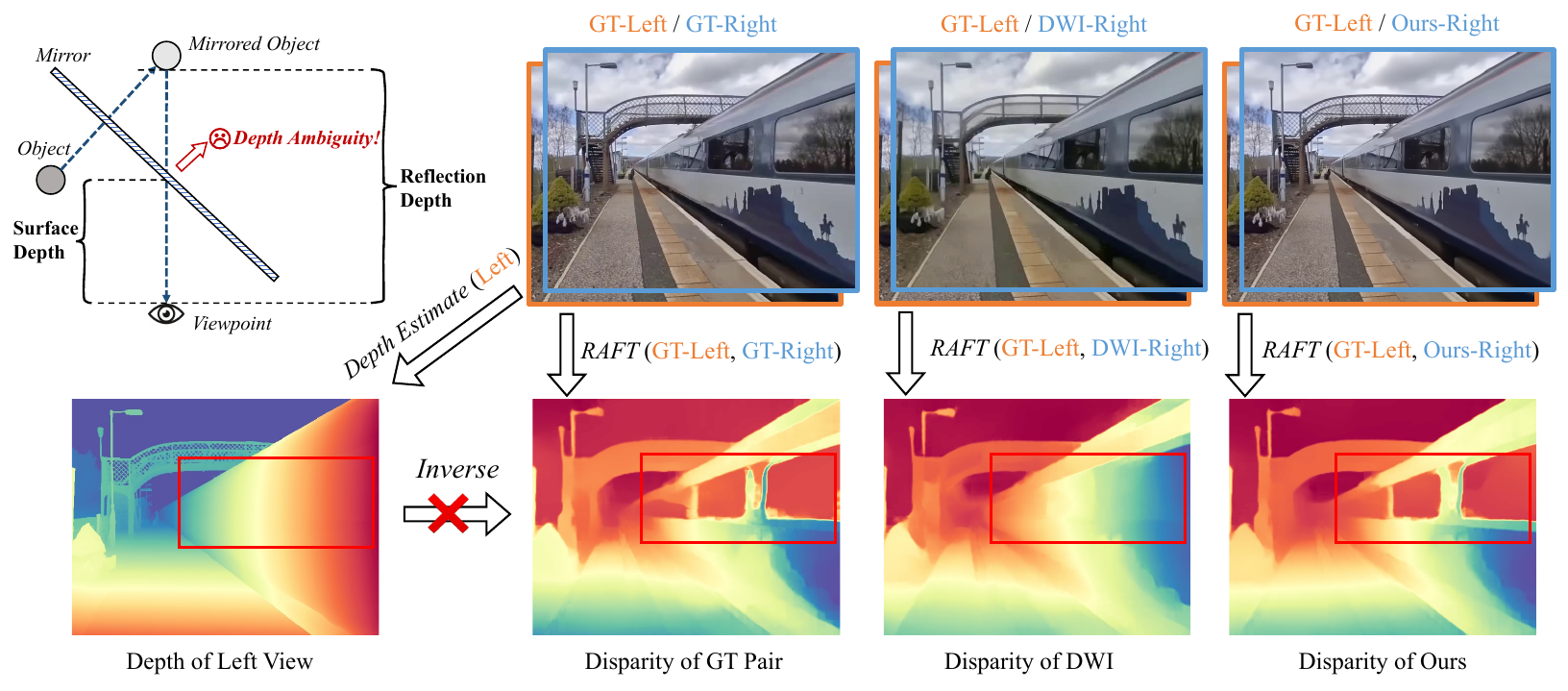}
    \caption{\textbf{Depth Ambiguity Issue}. As shown in the legend in the upper left corner of the figure, when there are specular reflections, there will be two depths at the mirror position: the depth of the mirror surface $d_S$ and the depth of the object's reflection $d_R$. In the real physical world, these two points are warped separately according to their respective disentangled depths. However, depth estimation algorithms cannot predict multiple depths at the same position. Therefore,  the inverse relationship between depth and disparity breaks down. This will cause Depth-Warp-Inpaint (DWI) type methods to predict results with incorrect disparity.}
    \label{fig:depth_embiguity}
\end{figure}

The rapidly improving stereoscopic displays, including virtual reality (VR) headsets, smart glasses, and 3D cinemas, have driven a significant and growing demand for immersive stereo video content. 
This technology offers a more immersive user experience compared to traditional 2D media, with widespread applications in the film, gaming, and education industries. 
Despite its potential, generating high-quality stereo content remains a formidable challenge.

Current immersive stereo video creation is constrained by two primary factors. First, native stereo video production necessitates specialized and expensive stereoscopic camera equipment coupled with complex post-processing workflows. Second, converting the vast repository of existing 2D video into a stereo format is a non-trivial task. 
For instance, the renowned manual conversion of the film \textit{Titanic} required a team of 300 engineers, took 60 weeks to complete, and resulted in a cost of 18 million dollars. 
The prohibitive costs and labor associated with these methods underscore the urgent need for robust and efficient automatic stereo video conversion techniques.

Recent advancements in video generation~\cite{blattmann2023stable, wan2025wan, yang2024cogvideox, chen2024videocrafter2} and depth estimation~\cite{yang2024depth, yang2024depthv2, ke2024repurposing, he2024lotus} have propelled significant progress in automated stereo video conversion. 
A predominant category of these approaches~\cite{zhao2024stereocrafter,shvetsova2025m2svid, dai2024svg, yu2025mono2stereo, lv2025spatialdreamer} adopts a multi-stage ``Depth-Warp-Inpaint'' (DWI) pipeline. 
This involves first estimating depth from the monocular video, then warping the image based on this depth map to create a second viewpoint, and finally inpainting any resulting occlusions. 
However, this paradigm suffers from a critical limitation in its sequential architecture: it creates a strong dependency on the initial depth estimation, where inaccuracies from the depth predictor propagate and compound through the pipeline, leading to substantial degradation in the final synthesized view.

\begin{figure}[t]
  \flushright 
  \centering
  \includegraphics[width=0.5\textwidth]{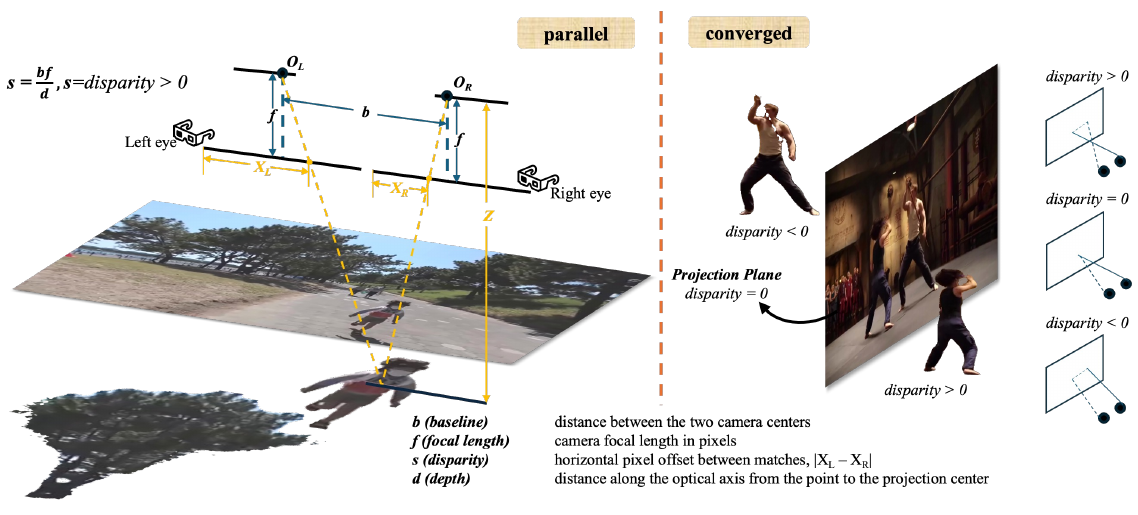}
  \caption{\textbf{Parallel vs. Converged.}
    In the parallel setup, when both eyes observe the same subject, the projected image points on the left and right views are denoted as $\mathbf{X_L}$ and $\mathbf{X_R}$, and their absolute difference $|\mathbf{X_L} - \mathbf{X_R}|$ defines the disparity $\boldsymbol{s}$.
    According to geometric relationships derived from similar triangles, $\boldsymbol{b}$, $\boldsymbol{f}$, $\boldsymbol{d}$, and $\boldsymbol{s}$ satisfy an inverse proportionality between disparity and depth when the baseline $\boldsymbol{b}$ and focal length $\boldsymbol{f}$ remain constant.
    In the converged configuration, a \textit{Zero-disparity Projection Plane} is present—objects in front of this plane yield positive disparity, while those behind it produce negative disparity.}
  \label{fig:paraVSconverge}
\end{figure}

More fundamentally, the depth-based approach suffers from inherent conceptual limitations. 
First, the depth-warp mechanism cannot adequately resolve scenes with depth ambiguity. 
As demonstrated in Figure \ref{fig:depth_embiguity}, in scenarios involving reflective surfaces such as mirrors, a single source pixel may correspond to multiple depth values---both the depth of the reflective surface itself and the depth of the reflected object.
A conventional depth-warp operation cannot model this one-to-many mapping, causing incorrect disparity in the synthesized view.
Second, the warping stage relies on a simple inverse relationship between depth and disparity, which is only valid for parallel camera setups, as shown in Figure~\ref{fig:paraVSconverge}.
However, this assumption breaks down for converged (toe-in) configurations---the standard for 3D cinema~\cite{yamanoue2006differences, wright2011parallel}---where the geometric relationship becomes more complex.


This geometric limitation highlights a broader issue: the diversity of stereo video formats and the lack of unified treatment in existing methods.
Based on our analysis of prior methods and a review of 3D film industry literature~\cite{yamanoue2006differences, wright2011parallel, fiveable2024interaxial, Seymour2012, vetro2008toward}, we identify two primary categories of stereo video with distinct characteristics, shown in Figure~\ref{fig:paraVSconverge}.
The parallel format uses cameras with parallel axes, establishing a simple inverse relationship between disparity and depth~\cite{birchfield1999depth, chang2018pyramid, wen2025foundationstereo, jing2024match,karaev2023dynamicstereo}.
In contrast, the converged format employs cameras in a ``toe-in'' configuration, creating a zero-disparity plane that is standard for 3D cinema~\cite{yamanoue2006differences, wright2011parallel}.
Several DWI methods assume parallel geometry, yet training data is in converged format~\cite{yu2025mono2stereo, zhao2024stereocrafter}, creating a fundamental mismatch.
Moreover, previous methods~\cite{zhao2024stereocrafter, yu2025mono2stereo, xie2016deep3d, lv2025spatialdreamer, geyer2025eye2eye,shvetsova2025m2svid,dai2024svg} were mostly trained on private data belonging to only one format, without explicitly distinguishing or providing unified treatment for both formats.
This has led to two critical consequences: first, the applicability of each method remains unclear; second, inappropriate evaluation protocols have emerged, such as benchmarking a model trained on parallel data against one trained on converged data~\cite{shvetsova2025m2svid, geyer2025eye2eye}, resulting in unfair comparisons.
To address this critical gap, we introduce UniStereo, the first large-scale, unified dataset for stereo video conversion that incorporates both parallel and converged stereo data. 
Our dataset comprises approximately 103,000 stereo video pairs with corresponding textual captions for training, along with a dedicated test set for standardized evaluation. 
We will release UniStereo publicly, hoping to contribute to the development of stereo video conversion.

To address these challenges, we introduce StereoPilot, an efficient model for high-quality stereo video conversion built upon our UniStereo dataset. 
To overcome error propagation in multi-stage pipelines, StereoPilot adopts an end-to-end architecture that directly synthesizes the novel view.
Instead of relying on explicit depth maps that struggle with ambiguous scenes, our approach leverages the rich generative priors from a pretrained video diffusion transformer.
Rather than depending on format-specific geometric assumptions, our method employs a data-driven approach that adapts to diverse stereo formats. We adopt a feed-forward architecture that performs direct regression to the target view, avoiding the randomness and computational overhead of iterative sampling while ensuring consistency and efficiency.
To handle both formats within a unified framework, we integrate a learnable domain switcher and a cycle consistency loss to ensure precise view alignment.
Comprehensive quantitative and qualitative experiments demonstrate that StereoPilot significantly outperforms state-of-the-art methods in both visual fidelity and processing speed.

In summary, our contributions are listed as follows:
\begin{itemize}
    \item We introduce UniStereo, the first large-scale, unified dataset for stereo video conversion, featuring both parallel and converged formats to enable fair benchmarking and model comparisons.
    
    \item We propose StereoPilot, an efficient feed-forward architecture that leverages a pretrained video diffusion transformer to directly synthesize the novel view. It overcomes the limitations of ``Depth-Warp-Inpaint'' methods (error propagation, depth ambiguity, and format-specific assumptions) without iterative denoising overhead, while integrating a domain switcher and cycle consistency loss for robust multi-format processing.
    
    \item Extensive experiments show StereoPilot significantly outperforms state-of-the-art methods on our UniStereo benchmark in both visual quality and efficiency.
\end{itemize}


\section{Related Works}
\label{sec:related work}

\subsection{Novel View Synthesis}
Novel view synthesis is a long-standing problem in computer vision and graphics. With advancements in 3D representation, such as Nerual Radiance Fields~\cite{mildenhall2021nerf} and 3D Gaussian Splatting~\cite{kerbl3Dgaussians}, static multi-view optimization-based novel view synthesis~\cite{ge2023ref, verbin2022ref, barron2021mip, barron2022mipnerf360, liu2023nero, zhang2020nerf++} has made significant progress. However, these methods typically require per-scene optimization with static, calibrated, multi-view dense images, which is computationally expensive, lacks generalization, and can only handle static synthesis.

To address the issues of generalization and the need of multi-view dense images, recent methods have explored camera-controlled video generation~\cite{he2024cameractrl, bahmani2024vd3d, bahmani2024ac3d, yu2024viewcrafter, ren2025gen3c, gao2024cat3d, wang2024motionctrl, hu2025ex, sun2024dimensionx} for single-image generalizable novel view synthesis. Given a target camera trajectory and a reference image, these methods can generate novel views that match the specified camera trajectory. More recent works~\cite{bai2025recammaster, wu2025cat4d, he2025cameractrl} focus on dynamic novel view synthesis. For example, ReCamMaster~\cite{bai2025recammaster} is able to reproduce the dynamic scene of an input video at novel camera trajectories. However, training such a video diffusion model requires multi-view calibrated videos, which are difficult to acquire in the real world. Moreover, the camera controllability is still limited.

In this work, we explore the task of stereo video synthesis, which is essentially a sub-problem of novel view synthesis with a fixed target camera. The training data can be more easily constructed using real-world videos with existing algorithms~\cite{jin2024stereo4d}. Since we focus on fixed novel-view synthesis, we do not explicitly require camera pose and can achieve better camera controllability.

\subsection{Stereo Video Generation}
Monocular-to-stereo video conversion, a specialized task within novel view synthesis, aims to generate a stereoscopic video given only the source viewpoint. Current approaches can be broadly classified into two families: multi-stage pipelines and end-to-end models.

\vspace{0.1cm}
\noindent \textbf{Multi-Stage Depth-Warp-Inpaint Pipelines}. The dominant paradigm follows a ``depth-warp-inpaint'' strategy~\cite{zhao2024stereocrafter, shvetsova2025m2svid, dai2024svg, yu2025mono2stereo, lv2025spatialdreamer}. These methods first estimate a per-pixel depth map from the monocular input video. This depth map is then used to explicitly warp the source view to the target view, which inevitably creates occlusions and artifacts. Finally, a generative model is employed to inpaint these missing regions. While popular, this multi-stage approach suffers from two fundamental limitations. First, the entire pipeline's performance is heavily dependent on the quality of the initial depth estimation. Inaccuracies in the predicted depth cascade through the subsequent stages, leading to severe geometric distortions and artifacts. Second, these methods are unable to handle scenes with depth ambiguity, such as those containing reflections or transparent materials, as shown in Figure~\ref{fig:depth_embiguity}. In these scenarios, a single pixel location corresponds to multiple depth values (e.g., the glass surface and the object behind it). By relying on a single depth value per pixel, this pipeline fails to model the correct physical optics, resulting in artifacts such as virtual images appearing ``baked'' onto the mirror surface. Third, warping relies on the assumption that depth and disparity follow an inverse relationship, which does not hold for converged camera configurations.

\begin{figure}[t]
  \centering
  \includegraphics[width=0.5\textwidth]{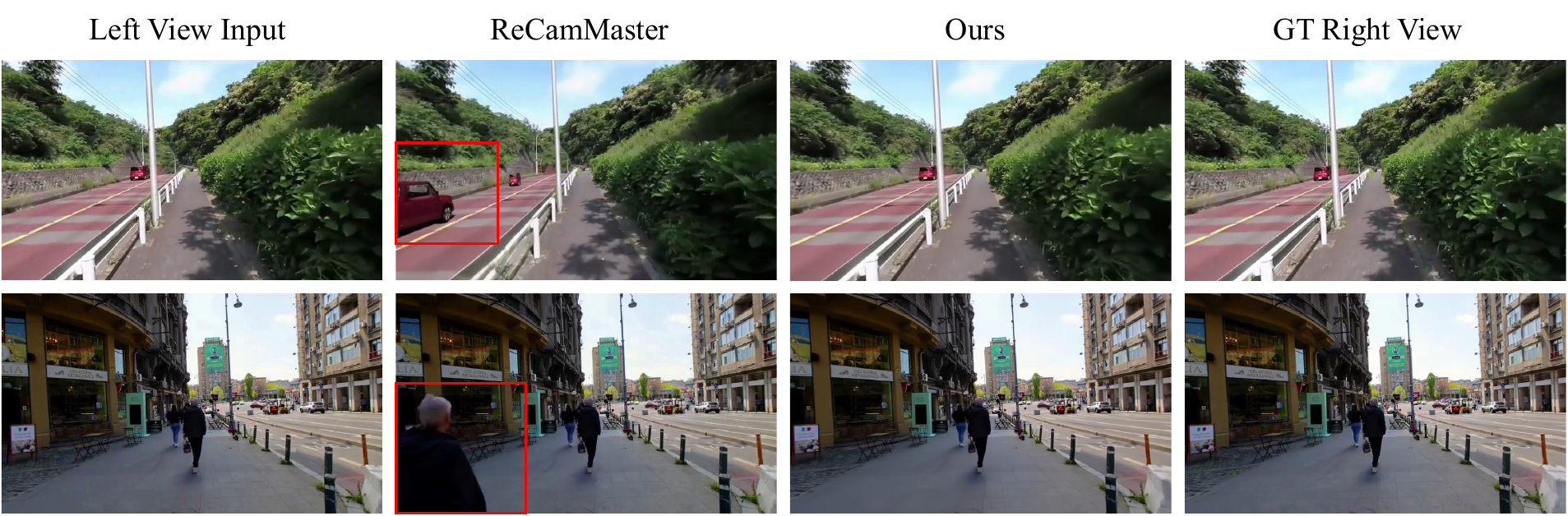}
  
  \caption{The inherent stochasticity of generative models can cause them to fabricate objects not present in the source view. As this figure illustrates, the right view generated by ReCamMaster erroneously introduces new artifacts, e.g., a car and a man (highlighted in red bounding box), that do not exist in the original input.}
  \label{fig:generation_issue}
\end{figure}

\vspace{0.1cm}
\noindent \textbf{End-to-End Synthesis Methods}. A second category of methods attempts to synthesize the target view directly in an end-to-end (E2E) fashion. Early E2E works, such as Deep3D~\cite{xie2016deep3d}, used CNNs to implicitly learn a soft disparity field for direct image prediction. However, it is trained on single images, failing to guarantee temporal consistency across video frames, and its CNN-based backbone limits scalability. More recently, methods like Eye2Eye~\cite{geyer2025eye2eye} have framed the task as conditional video generation using diffusion transformers. By iteratively denoising a noise-conditioned input view, the DiT can generate the other view directly. However, this iterative diffusion process requires dozens of sampling steps, making it computationally expensive and far too slow for practical applications. Moreover, due to the inherent stochasticity of video diffusion models, the diffusion process often produces hallucinated content that is misaligned with the source view, as shown in Figure~\ref{fig:generation_issue}.

Our method, StereoPilot, addresses these limitations. We propose a ``diffusion as feed-forward'' architecture that predict the target view in one step, efficiently leveraging diffusion priors while fundamentally avoiding the depth ambiguity problem. Unlike prior works trained on a single stereo format (converged or parallel), we introduce a learnable domain switcher that makes our model compatible with both, ensuring strong generalization. Finally, a cycle consistency loss is adopted to guide the model to generate the other view that aligns with the input view. StereoPilot thus achieves efficient, accurate, and robust mono-to-stereo conversion.

\section{Dataset Construction — UniStereo}



\begin{figure}[t]
  \centering
  \includegraphics[width=0.5\textwidth]{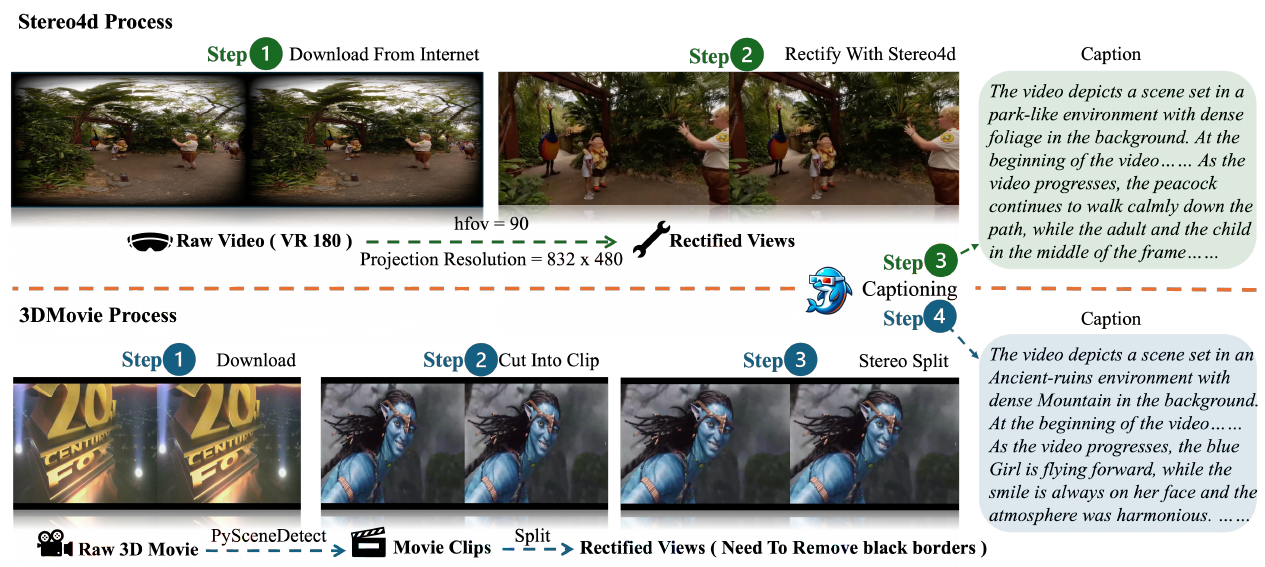}
  \caption{\textbf{UniStereo processing pipeline.} We use green icons with numbered steps to depict the \textbf{Stereo4D} pipeline: starting from the raw VR180 videos, we set hfov = 90° and specify the projection resolution to produce the final left- and right-eye monocular videos. Simultaneously, blue icons with numbered steps denote the \textbf{3DMovie} pipeline: we segment the source films into clips, filter out non-informative segments, convert from side-by-side (SBS) to left/right monocular views, and remove black borders. All resulting videos are captioned using ShareGPT4Video~\cite{chen2024sharegpt4video}.}
  \label{fig:UniStereo_process}
\end{figure}


As mentioned earlier, to unify the generation of both converged and parallel stereo data, and to address the issues of inappropriate evaluation protocols and unfair comparisons between models trained on parallel versus converged datasets, we propose UniStereo — the first large-scale unified dataset for 2D-to-3D video conversion that integrates both parallel and converged stereo data.

Specifically, UniStereo integrates two complementary components. The Stereo4D~\cite{jin2024stereo4d} subset provides large-scale parallel stereo pairs, as detailed in Section~\ref{subsec:Stereo4d}. To complement this, we construct 3DMovie, a new converged stereo video dataset derived from 3D cinematic productions, addressing the absence of publicly available converged-view data, as detailed in Section~\ref{subsec:3DMovie}. 

\subsection{Stereo4D}
\label{subsec:Stereo4d}
The Stereo4D subset serves as a large-scale collection of parallel stereo data designed to capture diverse real-world dynamics. It contains over 100K short clips depicting everyday scenes and activities, including both moving and static camera shots, sourced from approximately 7K online videos. Each clip spans 1–6 seconds and provides the camera-to-world (C2W) matrices for the left-eye view. The dataset contains a broad spectrum of indoor and outdoor environments and includes challenging scenes with reflective and transparent surfaces, ensuring rich visual diversity.

As illustrated in Figure~\ref{fig:UniStereo_process}, we follow the official preprocessing pipeline of Stereo4D~\cite{jin2024stereo4d} to generate consistent and rectified stereo pairs. Specifically, we convert VR180 videos with a horizontal field of view of 90° into rectified perspective videos at a resolution of 832 × 480. All videos are resampled to 16 fps and trimmed to a fixed length of 81 frames to ensure temporal uniformity across samples. Finally, we employ ShareGPT4Video~\cite{chen2024sharegpt4video} to automatically generate captions for each video, resulting in approximately 60K high-quality stereo video pairs.


\subsection{3DMovie}
\label{subsec:3DMovie}
To address the lack of publicly available converged-stereo datasets, we constructed a large-scale Converged 3D Movie Dataset (Figure~\ref{fig:UniStereo_process}). We curated 142 high quality 3D films and manually verified the consistency across both views to ensure valid stereo correspondence.

For temporal standardization, all movies were resampled to 16 fps. We then applied PySceneDetect~\cite{Castellano_PySceneDetect} for shot boundary detection, automatically segmenting each film into coherent stereo clips. Each clip was further divided into uniform segments of 81 frames to facilitate model training. Non-informative content, including opening logos, textual overlays, and ending credits, was removed to maintain visual relevance. After preprocessing, each video was converted from side-by-side (SBS) format into left–right paired monocular streams. To eliminate visual artifacts, we cropped edge black borders and resized all videos to 832 × 480 resolution. Finally, we employed ShareGPT4Video~\cite{chen2024sharegpt4video} to generate descriptive captions for every stereo pair. The resulting dataset contains approximately 48K high-quality stereo video pairs, providing a reliable foundation for research on converged-stereo understanding and generation.

\section{Method}

\begin{figure*}
    \centering
    \includegraphics[width=1.0\linewidth]{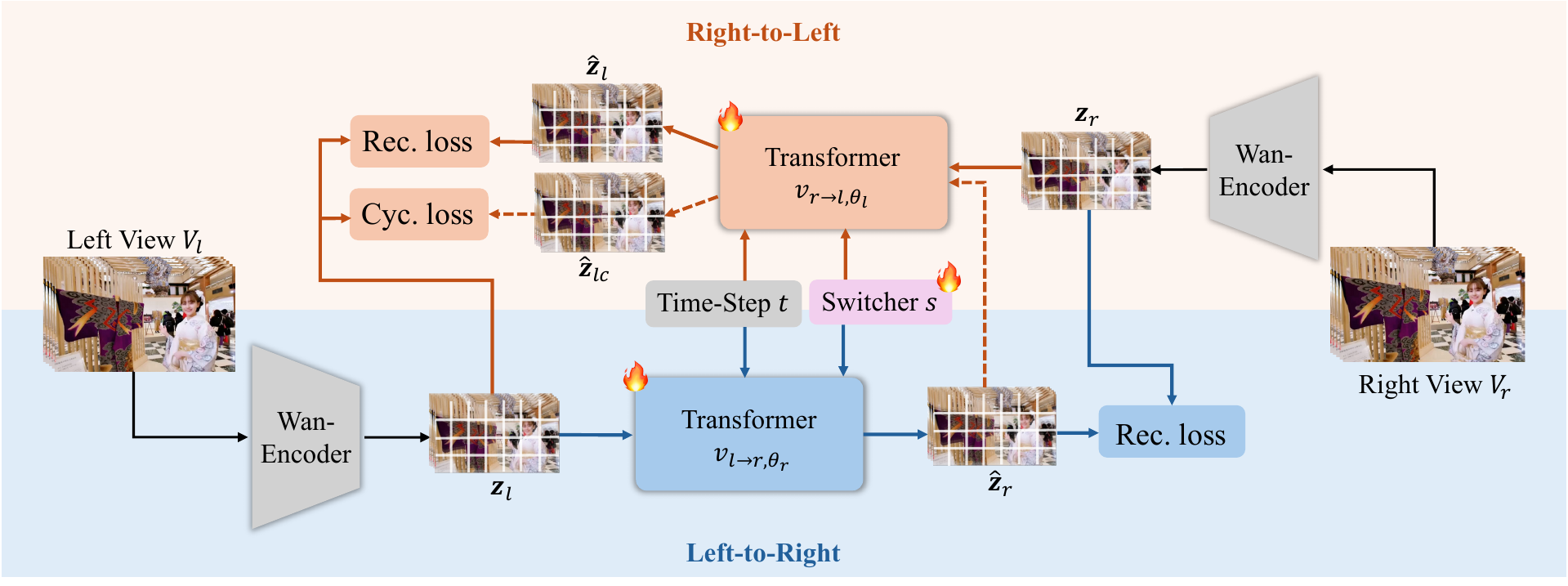}
    \caption{\textbf{The training framework of the proposed StereoPilot.} StereoPilot uses a \emph{single-step feed-forward} architecture (Diffusion as Feed-Forward) that incorporates a \emph{learnable domain switcher $s$} to unify conversion for both parallel and converged stereo formats. The entire model is optimized using a \emph{cycle-consistent training strategy}, combining reconstruction and cycle-consistency losses to ensure high fidelity and precise geometric alignment.
    The \textcolor[HTML]{205E9B}{\textbf{blue}} and \textcolor[HTML]{C04E15}{\textbf{orange}} lines represent the Left-to-Right and Right-to-Left reconstruction processes, and the \textcolor[HTML]{C04E15}{\textbf{orange dashed line}} denotes the $L \rightarrow R \rightarrow L$ cycle-consistency path.     }
    \label{fig:overall}
\end{figure*}

We begin with a brief overview of the conditional video diffusion models in Section \ref{subsec: preliminary}. Section \ref{subsec: diffasff} describes diffusion as feed-forward model in our method. We then introduce the domain switcher to include both converged and parallel formats in Section \ref{subsec: domainswitcher}, followed by the designed cycle consistent learning objective in Section \ref{subsec: cycleloss}. The overall is illustrated in Figure~\ref{fig:overall}.

\subsection{Preliminary}
\label{subsec: preliminary}
\noindent \textbf{Task Formulation.} The goal of monocular-to-stereoscopic video conversion is to synthesize a 3D stereoscopic video from a 2D video. We define this task as a conditional view synthesis problem. Given a ground-truth stereoscopic video pair consisting of a left view $V_l$ and a right view $V_r$, where $V_l, V_r \in \mathbb{R}^{N \times H \times W \times 3}$,  the objective is to predict one view conditioned on the other. Following prior works, we formulate our approach to take the left view video $V_l$ as input and the right view video $V_r$ as output. 

\vspace{0.1cm}
\noindent \textbf{Conditional Video Diffusion Models.} Recent advanced video generation models~\cite{wan2025wan, kong2024hunyuanvideo, KTeam2024Kling} are composed of a VAE and a Transformer network. The video diffusion transformers usually adopt Rectified Flow framework. It learns to map samples from a simple prior distribution to a complex data distribution (e.g., videos) by modeling the transport via an Ordinary Differential Equation (ODE). Let $\mathbf{z}_0 \sim p_{\text{data}}$ be a sample from the real data distribution and $\mathbf{z}_1 \sim p_1(\mathbf{z}) = \mathcal{N}(0, \mathbf{I})$ be a Gaussian noise. Rectified Flow defines a simple, straight-line path between these two points:
\begin{equation}
\mathbf{z}_t = (1-t) \mathbf{z}_0 + t \mathbf{z}_1, \quad t \in [0, 1].
\end{equation}
The velocity vector field along this path is constant and given by $v(\mathbf{z}_t) = \mathbf{z}_1 - \mathbf{z}_0$. The transformer network $v_\theta(\mathbf{z}_t, t, c)$ is then trained to approximate this vector field, often conditioned on context $c$ (e.g., a text embedding). The model is optimized using the flow-matching objective, which is an L2 regression loss:
\begin{equation}
    \mathcal{L}_{\text{FM}} = \mathbb{E}_{t \sim \mathcal{U}(0,1), \mathbf{z}_0, \mathbf{z}_1, c} \left[ \left\| v_\theta(\mathbf{z}_t, t, c) - (\mathbf{z}_1 - \mathbf{z}_0) \right\|_2^2 \right],
\end{equation}
where $t$ is sampled uniformly from $[0, 1]$. During inference, a sample is generated by first sampling noise $\mathbf{z}_0 \sim \mathcal{N}(0, \mathbf{I})$ and a condition $c$. The corresponding generative ODE is then solved from $t=1$ to $t=0$ using a numerical solver (e.g., Euler's method):
\begin{equation}
\mathbf{z}_{t_{\text{next}}} = \mathbf{z}_{t_{\text{curr}}} - \eta\cdot v_\theta(\mathbf{z}_{t_{\text{curr}}},t, c), 
\end{equation}
where $t_{\text{next}}< t_{\text{curr}}$ and $\eta$ ($0<\mathbf{\eta}\leq1$) denotes the step size, which is determined by the total number of inference time-steps. 

\vspace{0.1cm}
\noindent \textbf{Generative Novel View Video Synthesis.} To enable a text-to-video generation model to generate a novel view  $V_{target}$ conditioning on the input view $V_{source}$, some works try to inject the condition video $V_{source}$ to the generation process by concatenating $V_{source}$ with the noised $\mathbf{z}_t$, either on the channel dimension~\cite{geyer2025eye2eye} or the frame dimension~\cite{bai2025recammaster}. However, applying a generative paradigm to the mono-to-stereo task suffers from two significant limitations. First, a fundamental mismatch exists between stochastic generative models and the task's highly deterministic nature. Since occlusions between viewpoints typically constitute only a small fraction of the total image, the mapping for most pixels is deterministic. The model's inherent uncertainty is ill-suited for this, often introducing artifacts in non-occluded regions, as shown in Fig~\ref{fig:generation_issue}. Second, the iterative inference process incurs substantial computational overhead. Generating even a few seconds of video can require tens of minutes, restricting practical application in real-world scenarios.

\subsection{Diffusion as Feed-Forward Model}
\label{subsec: diffasff}
As we mentioned above, the stereo conversion task is highly deterministic, which contrasts with standard probabilistic diffusion models. To adapt a diffusion model for this task, we formulate it as a feed-forward network, which has shown success in some deterministic tasks like depth and normal estimation~\cite{he2024lotus,garcia2025fine}. We achieve this by modifying the standard training procedure: instead of sampling $t$ randomly, we fix the timestep to a small constant $t_0 = 0.001$. This value is chosen since the input state  at this near-zero timestep closely approximates our task's input data distribution, allowing us to effectively harness the model's generative priors. The process is thus reduced to a single-step deterministic prediction:
\begin{equation}
\mathbf{z}_r = v_\theta(\mathbf{z}_l, t_0, c) ,
\label{eq:feedforward}
\end{equation}
where $v_\theta$ predicts the right view $\mathbf{z}_r$ from the left view $\mathbf{z}_l$ and text conditioning $c$ in a single forward pass. This feed-forward methodology is computationally efficient (one inference step), and its deterministic nature fits the task. Furthermore, the pretrained knowledge in the video diffusion model also provides rich generation prior to completion of the occluded regions.

\subsection{Unified Conversion by Domain Switcher}
\label{subsec: domainswitcher}
Converged stereo data $D_c$ and parallel stereo data $D_p$ stereo data are distinct, and applications often require conversion between them. A naive approach trains two separate models, but this has key limitations. First, domain biases—such as parallel public datasets lacking synthetic content—cause models trained only on $D_p$ to fail on those styles. Second, separate training limits the data scale for each model, restricting generalization.

To address this, we propose a unified model that learns from both formats simultaneously. We use a conditional switcher, $s \in \{s_c, s_p\}$, to direct the model to predict either the converged or parallel target view. The switcher 
$s$ is a learnable one-dimensional vector that is added to the time embeddings of the diffusion model, enabling seamless domain switching without mutual interference. Formally, Equation. \ref{eq:feedforward} becomes
\begin{equation}
\mathbf{z}_r =
\begin{cases}
v_\theta(\mathbf{z}_l, t_0, c, s_c), & \text{if } (\mathbf{z}_l, \mathbf{z}_r) \in D_c \\
v_\theta(\mathbf{z}_l, t_0, c, s_p), & \text{if } (\mathbf{z}_l, \mathbf{z}_r) \in D_p
\end{cases}
.
\end{equation}
This unified strategy provides stronger generalization. It successfully resolves the challenge of processing synthetic animation styles in the parallel format, a key failure point for separately trained models.

\subsection{Cycle Consistent Loss}
\label{subsec: cycleloss}
Maintaining consistency between the left and right views is a critical objective in stereo video conversion. 
To improve the consistency, we introduce a cycle-consistency training objective, which is designed to enforce alignment between the generated target view and the source view. We leverage the inherent symmetry of the stereo conversion task, which can be formulated as both a left-to-right ($L \rightarrow R$) transformation and a right-to-left ($R \rightarrow L$) transformation. 
Our framework employs two distinct models: a generator $v_{l\rightarrow r, \theta_l}$ that translates the left view to the right, and a generator $v_{r\rightarrow l, \theta_r}$ that translates the right view to the left. Given a ground-truth stereo pair $(\mathbf{z}_l, \mathbf{z}_r)$, we first compute the synthesized target views $\hat{\mathbf{z}}_r$ (from left) and $\hat{\mathbf{z}}_l$ (from right):
\begin{equation}
\hat{\mathbf{z}}_r = v_{l\rightarrow r, \theta_l}(\mathbf{z}_l, t_0, c, s), \quad
\hat{\mathbf{z}}_l = v_{r\rightarrow l, \theta_r}(\mathbf{z}_r, t_0, c, s).
\end{equation}

\noindent The total training objective $\mathcal{L}$ combines two reconstruction losses with a cycle-consistency loss. The reconstruction losses ensure that the generated images are faithful to their respective ground-truth targets:
\begin{equation}
\mathcal{L}_{\text{recon}} = \|\hat{\mathbf{z}}_r - \mathbf{z}_r\|_2^2 + \|\hat{\mathbf{z}}_l - \mathbf{z}_l\|_2^2.
\end{equation}

\noindent The cycle-consistency loss enforces that if we translate a view to the other domain and back again, we should recover the original view. Here, we apply the $L \rightarrow R \rightarrow L$ cycle:
\begin{align}
 \mathcal{L}_{\text{cycle}} &=\|\mathbf{z}_l - \hat{\mathbf{z}}_{lc}\|_2^2, \\
    &= \|\mathbf{z}_l - v_{r\rightarrow l, \theta_r}(\hat{\mathbf{z}}_r, t_0, c, s)\|_2^2,
\end{align}
where $\hat{\mathbf{z}}_{lc}$ denotes the synthesized left view from the synthesized right view $\hat{\mathbf{z}}_r$. 
The final loss function is the sum of these components, which are jointly optimized:
\begin{equation}
\mathcal{L} = \mathcal{L}_{\text{recon}} + \lambda \cdot \mathcal{L}_{\text{cycle}}.
\end{equation}
\section{Experiments}
\subsection{Datasets and Evaluation Protocol}


\begin{figure*}
    \centering
    \includegraphics[width=1.0\linewidth]{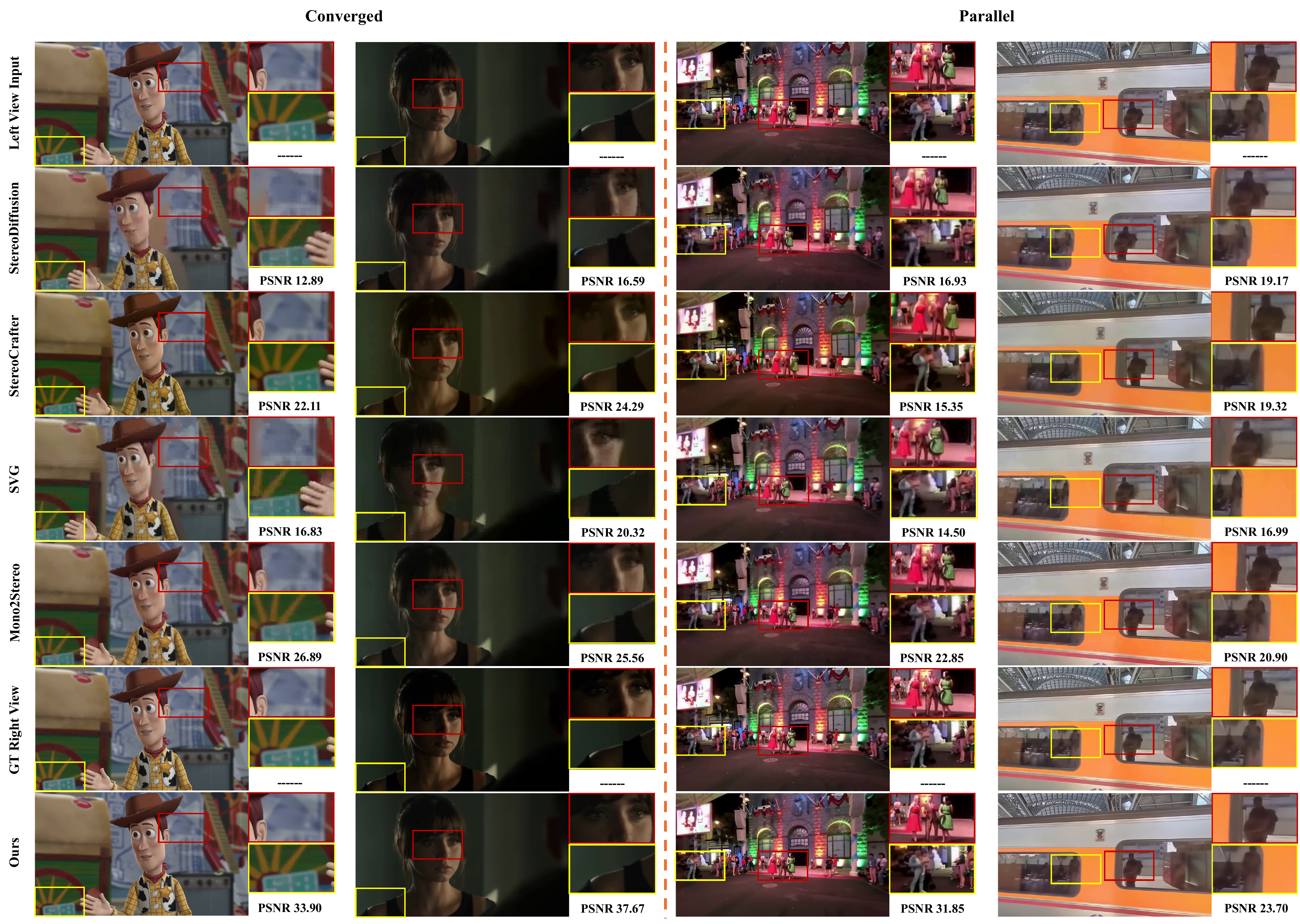}
    \caption{\textbf{Qualitative Results.} Our method achieves more accurate disparity estimation and preserves finer visual details on both Parallel and Converged data compared with existing baselines.}
    \label{fig:QuantitativeResults}
\end{figure*}

\vspace{0.1cm}
\noindent\textbf{Datasets.}
Our evaluation benchmark consists of two subsets: Stereo4D, which follows a parallel format, and 3DMovie, which follows a converged format.
We process all videos into 5-second clips sampled at 16 fps, resulting in a total of 81 frames per clip.

We extract 58,000 clips from the official training split of the Stereo4D dataset. For evaluation, and to ensure a fair comparison with previous methods, we follow prior work and randomly sample 400 video clips from the official Stereo4D test set~\cite{jin2024stereo4d}. For 3DMoive, we collect 44,879 video clips from a total of 126 3D movies for training. This collection is diverse, comprising 48 animated films and 78 live-action films. For the test set, we randomly sample 400 clips from 16 distinct 3D movies that are not included in the training set.

\vspace{0.1cm}
\noindent\textbf{Evaluation Protocol.}
We evaluate the performance of our method using metrics that assess both conversion quality and processing efficiency.
Unlike general visual generation tasks, our stereo conversion task provides corresponding ground-truth data, enabling direct measurement of the fidelity and alignment of the generated results.
Following prior work~\cite{shvetsova2025m2svid, wang2024stereodiffusion, yu2025mono2stereo}, we adopt a comprehensive set of widely used metrics to evaluate the quality of the generated stereo pairs, including PSNR, SSIM, MS-SSIM~\cite{zhang2018unreasonable}, and the perceptual metric LPIPS.
Furthermore, to better capture perceptual quality as judged by humans, we also include the SIOU metric, which was introduced in Mono2Stereo~\cite{yu2025mono2stereo} and has been shown to exhibit a strong correlation with human perception.
To measure computational performance, we report latency. All latency values presented in this paper refer to the time required to process a single 81-frame video clip (5 seconds at 16 fps) on a single GPU.

\subsection{Implementation Details} 
We use Wan2.1-1.3B~\cite{wan2025wan} as our backbone model.
During training, we adopt AdamW~\cite{loshchilov2017decoupled} to optimize the model parameters for approximately 30K iterations with a learning rate of $3\times10^{-4}$.
The coefficient $\lambda$ is set to 0.5 to ensure training stability.
Please refer to the Appendix for additional implementation details.

\begin{table*}[h]
\centering
\caption{Quantitative comparison on the Stereo4D and 3D Movie datasets. $\uparrow$ and $\downarrow$ indicate that higher is better and lower is better.}
\vspace{-0.2cm}
\label{tab:stereo_comparison_merged_reordered_correct}
\resizebox{\textwidth}{!}{
\begin{tabular}{ll ccccc ccccc l}
\toprule
\multirow{2}{*}{\textbf{Method}} & \multirow{2}{*}{\textbf{Venues}} & \multicolumn{5}{c}{\textbf{Stereo4D-Parallel Format}} & \multicolumn{5}{c}{\textbf{3D Movie-Converged Format}} & \multirow{2}{*}{\textbf{Latency}$\downarrow$} \\
\cmidrule(lr){3-7} \cmidrule(lr){8-12}
& & \textbf{SSIM}$\uparrow$ & \textbf{MS-SSIM}$\uparrow$ & \textbf{PSNR}$\uparrow$ & \textbf{LPIPS}$\downarrow$ & \textbf{SIOU}$\uparrow$ & \textbf{SSIM}$\uparrow$ & \textbf{MS-SSIM}$\uparrow$ & \textbf{PSNR}$\uparrow$ & \textbf{LPIPS}$\downarrow$ & \textbf{SIOU}$\uparrow$ & \\
\midrule
StereoDiffusion~\cite{wang2024stereodiffusion} & CVPR'24 & 0.642 & 0.711 & 20.541 & 0.245 & 0.252 & 0.678 & 0.612 & 20.695 & 0.341 & 0.181 & 60 min \\
StereoCrafter~\cite{zhao2024stereocrafter} & arXiv'24 & 0.553 & 0.562 & 17.673 & 0.298 & 0.226 & 0.706 & 0.799 & 23.794 & 0.203 & 0.213 & 1 min \\
SVG~\cite{dai2024svg} & ICLR'25 & 0.561 & 0.543 & 17.971 & 0.368 & 0.220 & 0.653 & 0.553 & 19.059 & 0.426 & 0.166 &  70 min \\
ReCamMaster~\cite{bai2025recammaster} & ICCV'25 & 0.542 & 0.525 & 17.229 & 0.312 & 0.239 & -- & -- & -- & -- & -- & 15 min \\
M2SVid~\cite{shvetsova2025m2svid} & 3DV'26 & -- & 0.915 & 26.200 & 0.180 & -- & -- & -- & -- & -- & -- & -- \\
Mono2Stereo~\cite{yu2025mono2stereo} & CVPR'25 & 0.649 & 0.721 & 20.894 & 0.222 & 0.241 & 0.795 & 0.810 & 25.756 & 0.191 & 0.201 & 15 min \\
\midrule
StereoPilot (Ours) & -- & \textbf{0.861} & \textbf{0.937} & \textbf{27.735} & \textbf{0.087} & \textbf{0.408} & \textbf{0.837} & \textbf{0.872} & \textbf{27.856} & \textbf{0.122} & \textbf{0.260} & \textbf{11 s} \\
\bottomrule
\end{tabular}
}
\label{tab:comparison_all} 
\end{table*}

\subsection{Main Results}

\noindent\textbf{Baselines.} 
 We compare StereoPilot with several state-of-the-art methods, including StereoDiffusion~\cite{wang2024stereodiffusion}, SVG~\cite{dai2024svg}, ReCamMaster~\cite{bai2025recammaster}, and Mono2Stereo~\cite{yu2025mono2stereo}.
In addition, we also include two influential works from the community, StereoCrafter~\cite{zhao2024stereocrafter} and M2SVid~\cite{shvetsova2025m2svid}, for comprehensive comparison.
We use the default parameter settings provided in each baseline method to ensure a fair comparison.

\begin{table}[h]
\centering
\caption{Ablation results of our proposed method}
\vspace{-0.2cm}
\label{tab:stereo_comparison_merged_reordered_correct}
\resizebox{\columnwidth}{!}{
\begin{tabular}{l ccccc}
\toprule
\textbf{Method} & \textbf{SSIM}$\uparrow$ & \textbf{MS-SSIM}$\uparrow$ & \textbf{PSNR}$\uparrow$ & \textbf{LPIPS}$\downarrow$ & \textbf{SIOU}$\uparrow$ \\
\midrule
Baseline & 0.833 & 0.891 & 26.954 & 0.143 & 0.319 \\
Baseline w/ switcher & 0.845 & 0.895 & 27.332 & 0.118 & 0.323 \\
Baseline w/ switcher + $\mathcal{L}_{\text{cycle}}$  & 0.849 & 0.905 & 27.796 & 0.105 & 0.334 \\
\bottomrule
\end{tabular}
}
\label{tab:comparison_ablation}
\end{table}

\begin{table}[h]
\centering
\caption{Performance on UE5 synthetic style parallel format data.}
\vspace{-0.2cm}
\label{tab:stereo_comparison_merged_reordered_correct}
\resizebox{\columnwidth}{!}{
\begin{tabular}{l ccccc}
\toprule
\textbf{Method} & \textbf{SSIM}$\uparrow$ & \textbf{MS-SSIM}$\uparrow$ & \textbf{PSNR}$\uparrow$ & \textbf{LPIPS}$\downarrow$ & \textbf{SIOU}$\uparrow$ \\
\midrule
Baseline & 0.791 & 0.881 & 28.377 & 0.150 & 0.291 \\
Baseline w/ switcher & 0.824 & 0.905 & 29.614 & 0.144 & 0.305 \\
\bottomrule
\end{tabular}
}
\label{tab:domain_bias}
\end{table}

\vspace{0.1cm}
\noindent\textbf{Quantitative Results.}
As demonstrated in Table \ref{tab:comparison_all}, our StereoPilot outperforms all the baselines in all 5 quantitative metrics. Furthermore, our method requires only 11 seconds to complete an 81-frame video's stereo conversion, showing significant superiority on computational efficiency. Note that we do not report results of M2SVid on 3DMovies, since code of M2SVid has not been released yet. We adopt the performance on 400 test samples on Stereo4D they reported in the paper, aligning with our test setting. For ReCamMaster, we first compare it on the Stereo4D, as Stereo4D provides ground truth camera tracks of the target view, which is necessary for ReCamMaster to generate the target view video. However, ReCamMaster shows weak camera control ability and tends to generate new objects that never appear in the source view. Considering ReCamMaster's performance on Stereo4D, which has an accurate target view camera track, was already poor, we did not further test it on 3DMovie. Because extracting the camera track for the target view's video is difficult and will also introduce additional cumulative errors, which have been discussed in \cite{luo2025camclonemaster}.

\vspace{0.1cm}
\noindent\textbf{Qualitative Results.}
As illustrated in Figure~\ref{fig:QuantitativeResults}, our method consistently outperforms all the baselines in both disparity accuracy and visual quality on the Converged and Parallel test sets. In the Converged cases, the baselines—StereoDiffusion, StereoCrafter, SVG, and Mono2Stereo—frequently produce noticeable disparity errors, which are clearly reflected in the highlighted regions. While Mono2Stereo performs comparatively better, it still exhibits misalignment with the ground truth and introduces visible artifacts. The remaining methods suffer from severe blur in repainted areas, with StereoCrafter additionally showing a persistent yellow color shift. In the second Converged example, facial and shoulder close-ups further reveal consistent disparity mistakes and detail loss among all baselines, resulting in noticeably blurrier outputs than ours. Similar observations hold for the Parallel set: all four baselines fail to predict correct disparity, and their reconstructed textures appear soft and lacking detail, whereas our method maintains sharpness and stable geometry. Finally, under challenging mirror conditions, all DWI-based approaches incorrectly estimate the reflected disparity, with StereoDiffusion and SVG even producing distortions. The PSNR annotations verify that our results remain significantly better aligned with the ground truth than all competing methods.

\subsection{Ablation Study}
\begin{figure}[t]
  \centering
  \includegraphics[width=0.5\textwidth]{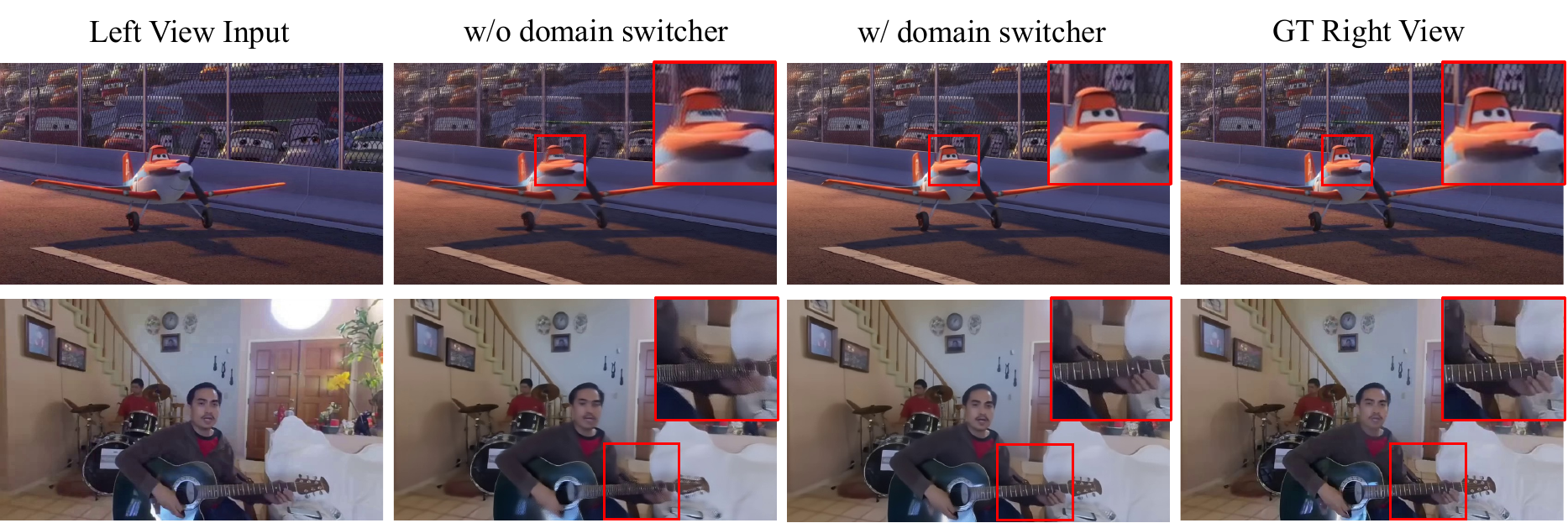}
  \caption{\textbf{Results w/ and w/o domain switcher.} Our proposed domain switcher enhances the model's generalization ability. }
  \label{fig:ablation_switcher}
\end{figure}

To validate our design within StereoPilot, we conduct ablation studies in this section. Detailed results are provided in Table \ref{tab:comparison_ablation}. Initially, we train two models in a Diffusion Feed-Forward manner on the Stereo4D and 3DMovie datasets, respectively, to establish our baseline. Then the domain switcher is injected to enable unified conversion and strong generalization performance, as shown in Figure~\ref{fig:ablation_switcher}. Finally, we ablate the cycle consistent loss on the model with the switcher. The findings from these ablations validate the effectiveness of our proposed module and training target, demonstrating that each design plays a vital role in optimizing the StereoPilot for stereo conversion tasks. Note that we average the performance in two datasets for simplicity.

Furthermore, in order to verify the effectiveness of the domain switcher in solving the domain bias in data distribution, that is, the model's generalization ability on parallel-style synthetic/anime-style stereo videos that have not appeared in the training set. We used Unreal Engine 5~\cite{UE5Website} to render 200 synthetic data style parallel 3D videos for testing. The performance improvement in Table \ref{tab:domain_bias} demonstrates the effectiveness of our designed domain switcher. Please refer to the Appendix for details.

\subsection{Conclusion and Limitation}
\noindent \textbf{Conclusion.} We introduce UniStereo, the first large, unified dataset with parallel and converged formats for standardized evaluation. We also propose StereoPilot, an efficient feed-forward model using a pretrained video diffusion transformer to directly synthesize the second view, avoiding ``Depth-Warp-Inpaint'' limitations like depth ambiguity. We integrate a domain switcher for format flexibility and a cycle consistency loss for high fidelity. Experiments show StereoPilot outperforms SOTA methods in visual quality and efficiency.

\vspace{0.1cm}
\noindent \textbf{Limitation.} Our model’s 11-second conversion time for a 5-second video is not yet sufficient for real-time applications such as live streaming. We plan to address this limitation in future work by exploring autoregressive conversion.

{
    \small
    \bibliographystyle{ieeenat_fullname}
    \bibliography{main}
}


\clearpage
\setcounter{page}{1}
\maketitlesupplementary

\section{Depth Ambiguity Details}
\label{sec:depth_ambiguity_details}
\begin{figure}[t]
  \centering
  \includegraphics[width=0.5\textwidth]{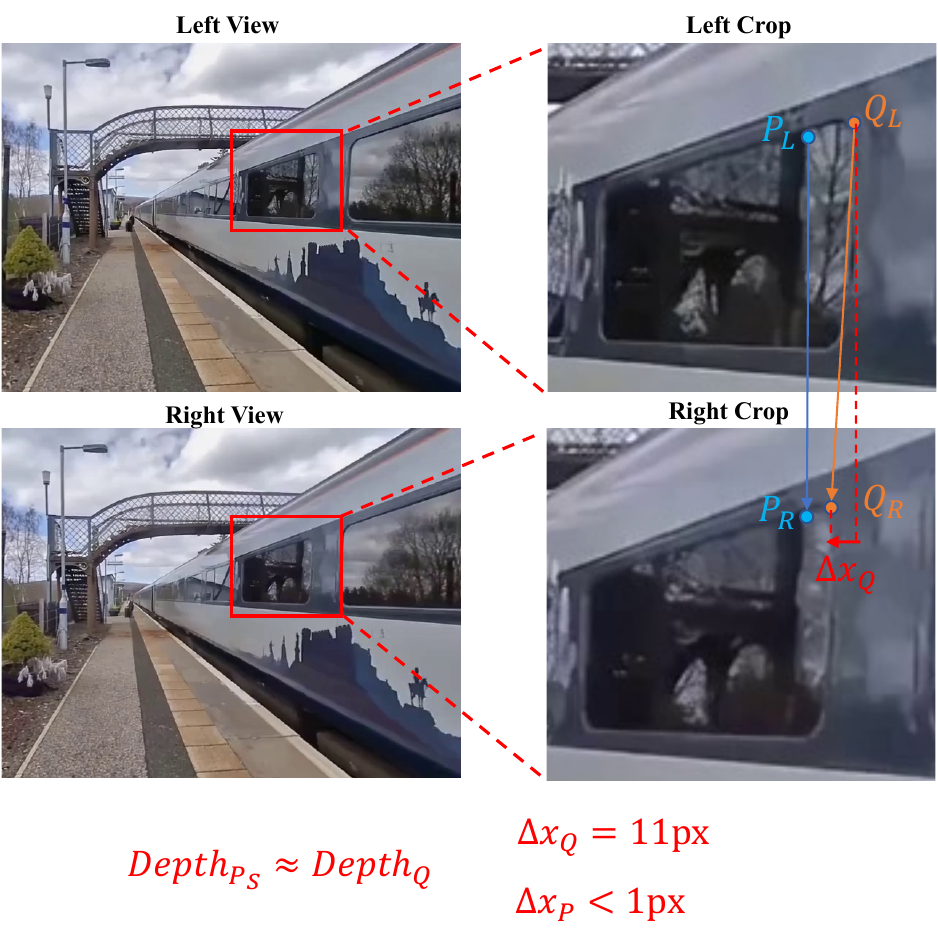}
  \caption{\textbf{Detailed Analysis of Depth Ambiguity. }}
  \label{fig_supp:depth_ambiguity_details}
\end{figure}

To demonstrate the challenge of depth ambiguity, we provide an in-depth analysis in Figure~\ref{fig_supp:depth_ambiguity_details} by tracking two representative keypoints across the stereo views. Let $P$ denote a pixel corresponding to a specular reflection (a utility pole) on the mirror surface, and $Q$ denote the position of the top-right outer frame outside the mirror.

While the physical depth of the mirror surface at $P$ ($D_{P_S}$) is proximate to the frame at $Q$ ($D_Q$), satisfying $D_{P_S} \approx D_Q$, their optical characteristics differ significantly. Specifically, the virtual imaging depth of the reflection at $P$ ($D_{P_R}$) is substantially larger than the physical depth of the frame ($D_{P_R} \gg D_Q$). Consequently, this discrepancy leads to distinct disparity behaviors: the reflected content at $P$ exhibits negligible parallax ($\Delta x_P \approx 0$), whereas the physical frame at $Q$ shows significant displacement ($\Delta x_Q = 11$ \text{px}). 

This manifests as a counter-intuitive visual phenomenon where the mirror surface and the reflected content appear disentangled: \textbf{\textit{in the transition between views, the mirror frame undergoes significant horizontal translation, while the specular reflection remains virtually stationary}}.


\section{UniStereo Construction Details}
\label{sec:unistereo_details}
We provide a detailed description of the data construction pipelines for both the Stereo4D (Section~\ref{sec:Stereo4d_details}) and 3DMovie (Section~\ref{sec:3DMovie_details}) datasets, outlining each processing step from raw videos to the final training-ready stereo pairs.

\subsection{Stereo4D Construction Details}
\label{sec:Stereo4d_details}
\begin{figure*}
    \centering
    \includegraphics[width=1.0\linewidth]{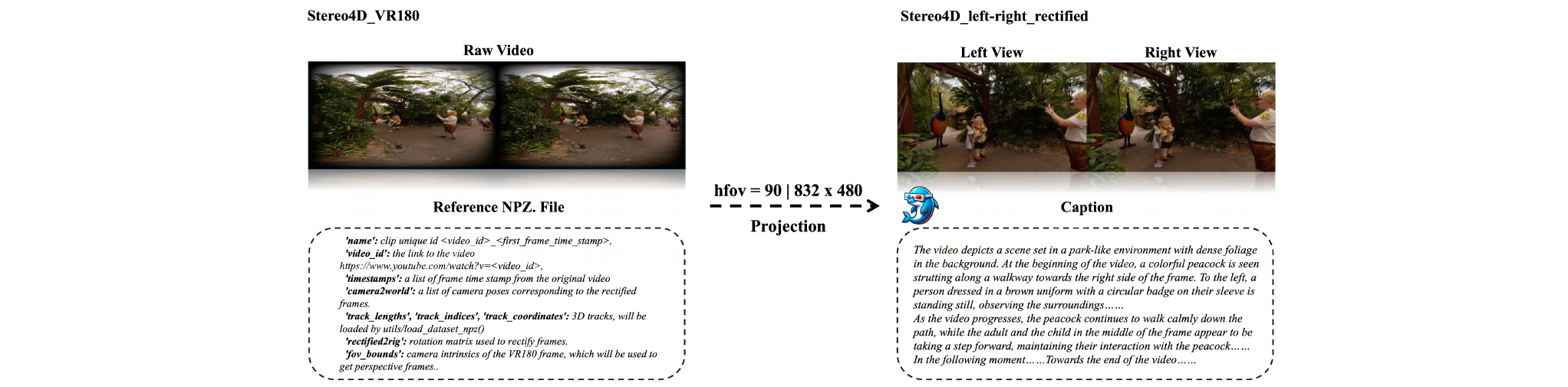}
    \caption{\textbf{Stereo4D processing pipeline.} The detailed process of Stereo4D data processing.}
    \label{fig:Stereo4d_process}
\end{figure*}

As illustrated in Figure~\ref{fig:Stereo4d_process}, we now describe the construction pipeline for the Stereo4D subset. We start from the official Stereo4D release~\cite{jin2024stereo4d}, which defines a corpus that links to about 4~TB of raw VR180 videos on YouTube and provides about 4~TB of accompanying NPZ metadata. The raw videos are VR180 clips on YouTube~\cite{youtube} that we download using yt-dlp~\cite{yt_dlp} together with the Stereo4D downloader~\cite{stereo4d_downloader}, resulting in roughly 7K source videos. Downloading and uploading this volume of data on a single machine typically takes three to four weeks. The NPZ files record, for each video, its video\_id for downloading, as well as the rectification parameters and camera2world matrices, which we later use as camera pose inputs when running ReCamMaster~\cite{bai2025recammaster} for inference. We obtain these NPZ files by following the instructions provided by Stereo4D~\cite{jin2024stereo4d}.

With the raw videos and metadata in place, we then follow the rectification pipeline specified in Stereo4D. Concretely, we set the output resolution in the rectification code to 832 × 480 and the horizontal field-of-view (hfov) to $90^\circ$ to convert each VR180 video into a pair of per-eye monocular videos. We run this rectification process on a cluster of 32 machines, each equipped with 32 CPU cores, for about two to three weeks, which yields on the order of 100K clips with durations between 1 and 6 seconds. To make the data compatible with Wan2.1-1.3B~\cite{wan2025wan} and to standardize the training inputs, we then resample all clips to 16~fps and retain only those whose length is at least 81 frames. From these filtered clips, we further process each one into an 81-frame sequence at a fixed resolution of 832 × 480. Finally, we apply ShareGPT4Video~\cite{chen2024sharegpt4video} to generate captions for the left-eye video of each stereo pair, obtaining approximately 60K stereo video–caption pairs for training and evaluation.

\begin{figure*}
    \centering
    \includegraphics[width=1.0\linewidth]{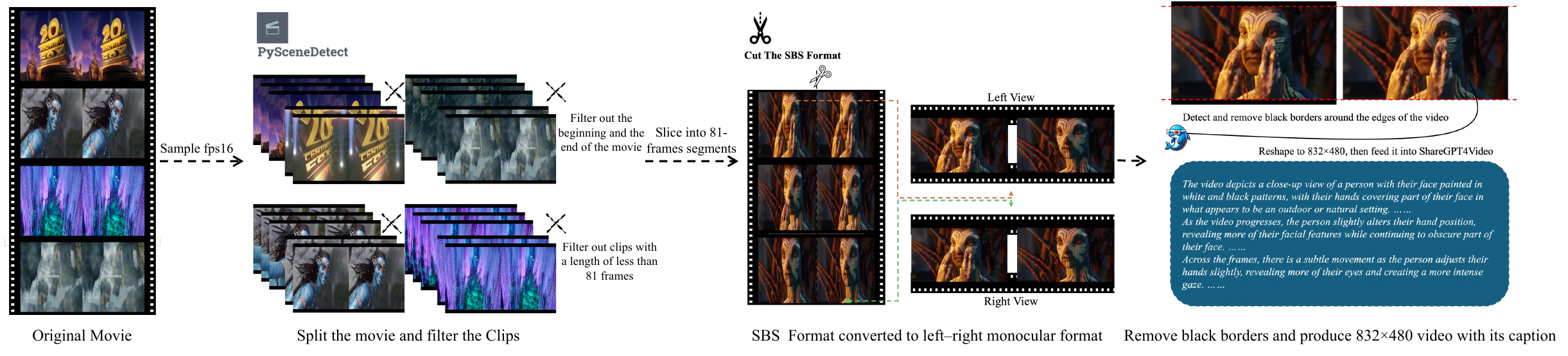}
    \caption{\textbf{3DMovie processing pipeline.} The detailed process of 3DMovie data processing.}
    \label{fig:3DMovie_process}
\end{figure*}

\subsection{3DMovie Construction Details}
\label{sec:3DMovie_details}
As illustrated in Figure~\ref{fig:3DMovie_process}, we now describe the construction pipeline of the 3DMovie dataset in detail. We start by collecting 159 convergent 3D movies and manually inspecting the stereo layout of each title. We retain only those movies encoded in left–right side-by-side (SBS) format and discard those using top–bottom layouts, so that all data share a consistent arrangement. Next, we examine the disparity pattern of every remaining movie. We find that some titles do not follow a true convergent stereo setup: switching between the left and right views only produces a horizontal shift in the image, rather than the slight rotational change typical of convergent cameras with a well-defined zero-disparity plane. Because such “pseudo-stereo” content may introduce harmful biases into model training, we remove these movies and finally keep 142 high-quality convergent 3D films.

We then standardize the data format. We first resample each movie to 16 fps, so that all videos share a unified temporal resolution that satisfies the stability requirements of the Wan2.1-1.3B~\cite{wan2025wan} model. Building on this, we apply PySceneDetect~\cite{Castellano_PySceneDetect} to segment each movie into short clips, ensuring that frames within the same clip mostly belong to a single scene and therefore exhibit a relatively consistent disparity pattern, which in turn makes stereo learning easier and accelerates training convergence. These scene clips have varying durations, so for training convenience and to match the temporal window of our model, we discard clips shorter than 81 frames. In addition, we trim the beginning and end of each movie to remove segments dominated by credits and production subtitles, which do not provide useful visual content. From this refined pool of clips, we generate fixed-length training samples by further slicing them into 81-frame segments. When a single clip yields multiple 81-frame segments, we index them in temporal order and retain only the segments with odd indices, thereby reducing redundancy while preserving temporal diversity.

Next, we process each 81-frame SBS segment into per-eye videos. We first split the SBS format into separate monocular sequences for the left and right eyes, and at the same time restore the original per-eye width so that each view has the correct spatial resolution. Some movies contain black borders around the image, which would otherwise introduce artificial structures into the data. To remove them in a consistent manner, we detect black borders on the left-eye video and then apply the same symmetric cropping to both left and right views. After this step, each retained segment becomes a pair of 81-frame, border-free left/right monocular videos. We then adjust the spatial resolution to match the input requirements of Wan2.1-1.3B~\cite{wan2025wan}. Based on the aspect ratio, we select video pairs whose size is close to 832 × 480 and reshape them to exactly 832 × 480. This results in a collection of stereo video pairs, each with 81 frames at a resolution of 832 × 480. Finally, we use ShareGPT4Video~\cite{chen2024sharegpt4video} to generate captions for the left-eye video of each pair, yielding approximately 48K high-quality stereo video–caption pairs for training and evaluation.

\section{UE5 Synthetic Dataset}
\label{sec:ue5_data_details}

\begin{figure}[t]
  \centering
  \includegraphics[width=0.5\textwidth]{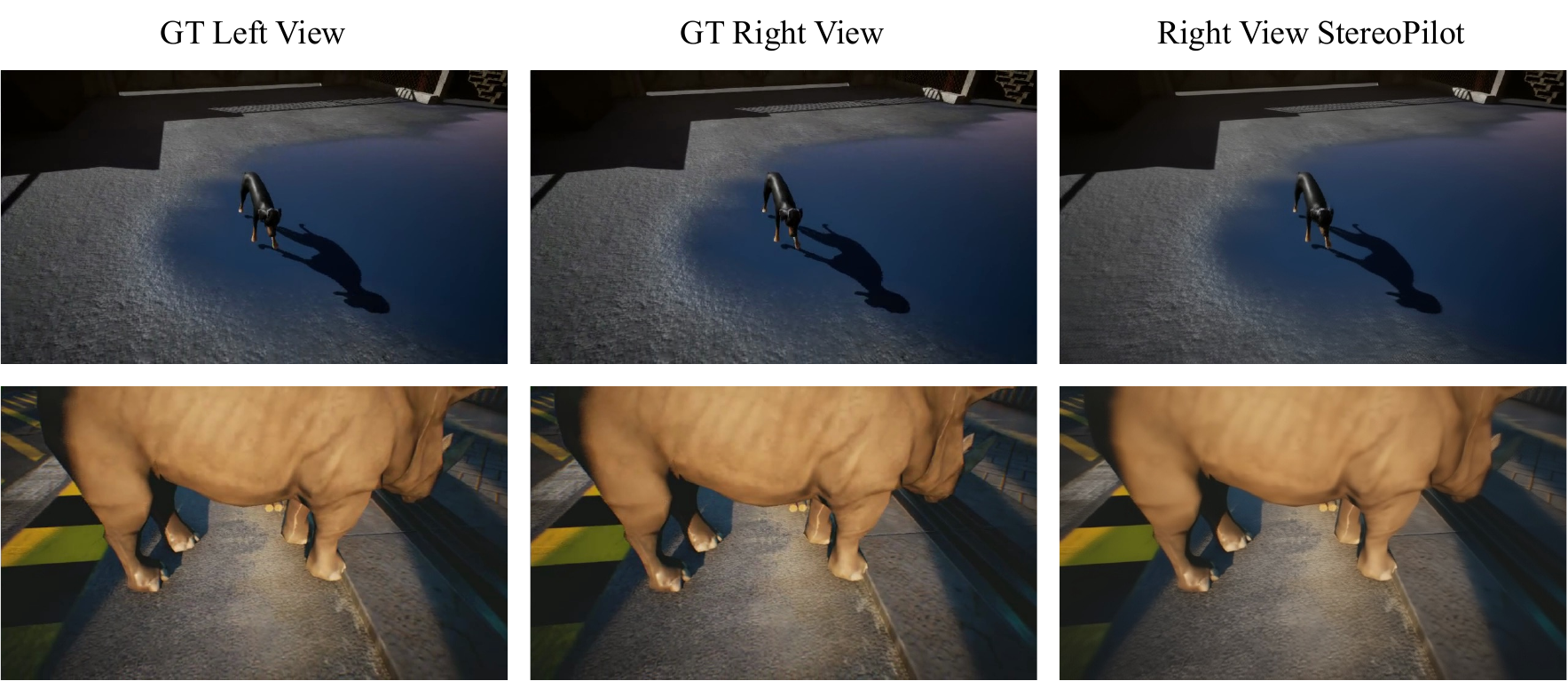}
  \caption{\textbf{Examples of UE5-Rendered Stereo Video Data.}}
  \label{fig_supp:ue_cases}
\end{figure}

To validate the generalization on the synthetic style paralleled format stereo video of StereoPilot, we construct a synthetic benchmark based on 6 randomly selected Unreal Engine~\cite{UE5Website} scenes under both daytime and nighttime conditions. To approximate human binocular vision and obtain paired stereo videos, we place two virtual cameras with a fixed baseline of 6.3 cm~\cite{jin2024stereo4d}. The scenes feature 28 categories of dynamically animated animals as the primary foreground subjects, and in total, we render 200 stereo video pairs, which are used to rigorously evaluate the effectiveness of our proposed model. As shown in Figure~\ref{fig_supp:ue_cases}, our model still demonstrates stable generalization ability on synthetic style parallel stereo datasets.

\section{Geometry Relationship in Stereo Vision}
\label{sec:geo_relation}

In this section, we first define the following common parameters for the stereo rig in stereoscopic vision for both parallel and converged camera configuration:

\begin{itemize}
    \item \textbf{Baseline ($B$)}: The distance between the optical centers of the left ($O_L$) and right ($O_R$) cameras.
    \item \textbf{Focal Length ($f$)}: The distance from the optical center to the image plane (assumed identical for both cameras).
    \item \textbf{Point $P(X, Y, Z)$}: A target point in 3D space located at depth $Z$.
    \item \textbf{Projections ($x_l, x_r$)}: The horizontal coordinates of point $P$ projected onto the left and right image planes, respectively.
\end{itemize}

\subsection{Parallel Stereo Video}
\label{subsec:parallel_stereo}

In a parallel stereo configuration, the two cameras are aligned such that their optical axes are strictly parallel. This setup is mathematically equivalent to standard epipolar geometry.

\subsubsection{Visual Representation}
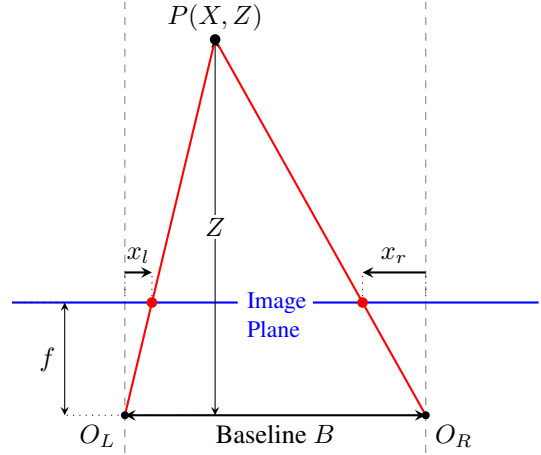
\begin{figure}[t]
\centering
    \begin{tikzpicture}[scale=1.0, >=stealth]
        \def\f{1.5}      
        \def\B{4.0}      
        \def\Z{5.0}      
        \def\X{1.2}      

        \coordinate (OL) at (0,0);        
        \coordinate (OR) at (\B,0);       
        \coordinate (P) at (\X, \Z);      
        
        \draw[dashed, gray] (0, -0.5) -- (0, \Z+0.5);
        \draw[dashed, gray] (\B, -0.5) -- (\B, \Z+0.5);

        \draw[thick, blue] (-1.5, \f) -- (1.5, \f);
        \draw[thick, blue] (\B-1.5, \f) -- (\B+1.5, \f);
        \node[blue, right, font=\small] at (1.5, \f) {Image};
        \node[blue, right, font=\small] at (1.5, \f-0.35) {Plane};

        \draw[thick, red] (OL) -- (P);
        \draw[thick, red] (OR) -- (P);

        \pgfmathsetmacro{\xl}{\f * \X / \Z}
        \coordinate (PL) at (\xl, \f);
        
        \pgfmathsetmacro{\xrLocal}{\f * (\X - \B) / \Z}
        \coordinate (PR) at (\B + \xrLocal, \f);

        \fill[red] (PL) circle (2pt);
        \fill[red] (PR) circle (2pt);

        \draw[<->, thick] (OL) -- (OR) node[midway, below] {Baseline $B$};
        \draw[<->] (-0.8, 0) -- (-0.8, \f) node[midway, left] {$f$};
        \draw[dotted] (OL) -- (-0.8, 0);
        \draw[dotted] (-1.5, \f) -- (-0.8, \f);

        \draw[<->] (\X, 0) -- (\X, \Z) node[midway, fill=white, inner sep=1pt] {$Z$};
        \draw[dotted] (\X, 0) -- (OL); 

        \draw[->, thick] (0, \f+0.4) -- (\xl, \f+0.4) node[midway, above] {$x_l$};
        \draw[dotted] (0, \f) -- (0, \f+0.4);
        \draw[dotted] (PL) -- (PL |- 0, \f+0.4);

        \draw[->, thick] (\B, \f+0.4) -- (\B + \xrLocal, \f+0.4) node[midway, above] {$x_r$};
        \draw[dotted] (\B, \f) -- (\B, \f+0.4);
        \draw[dotted] (PR) -- (PR |- 0, \f+0.4);

        \fill (OL) circle (1.5pt) node[below left] {$O_L$};
        \fill (OR) circle (1.5pt) node[below right] {$O_R$};
        \fill (P) circle (2pt) node[above] {$P(X,Z)$};
    \end{tikzpicture}
    \caption{Geometry of parallel stereo vision. By convention, $x_l$ is positive and $x_r$ is negative for a point located between the camera axes.}
    \label{fig:parallel_geometry}
\end{figure}

The geometry is illustrated in Figure~\ref{fig:parallel_geometry}. We assume the world coordinate system origin is located at the optical center of the left camera, $O_L = (0,0,0)$.

\subsubsection{Mathematical Derivation}

Using similar triangles, we relate the physical coordinates to the image plane coordinates.

\paragraph{Left Camera:}
With the camera at the origin $(0,0,0)$, the projection is:
\begin{equation}
    x_l = \frac{f \cdot X}{Z}
    \label{eq:left_proj}
\end{equation}

\paragraph{Right Camera:}
With the camera shifted by $B$ along the X-axis, the local X-coordinate is $(X - B)$. The projection is:
\begin{equation}
    x_r = \frac{f \cdot (X - B)}{Z}
    \label{eq:right_proj}
\end{equation}

\paragraph{Disparity and Depth:}
Disparity $d = x_l - x_r$ is calculated as:
\begin{equation}
\begin{split}
    d &= \frac{f X}{Z} - \frac{f (X - B)}{Z} \\
    d &= \frac{f \cdot B}{Z}
\end{split}
\end{equation}

Rearranging for depth $Z$:
\begin{equation}
    Z = \frac{f \cdot B}{d}
\end{equation}

\subsection{Converged Stereo Video}
\label{subsec:converged_stereo}

In stereoscopic 3D video creation, a \textit{toed-in} (or converged) configuration is often used. The optical axes of the two cameras are rotated to intersect at a specific convergence distance $Z_c$.

\subsubsection{Visual Representation}
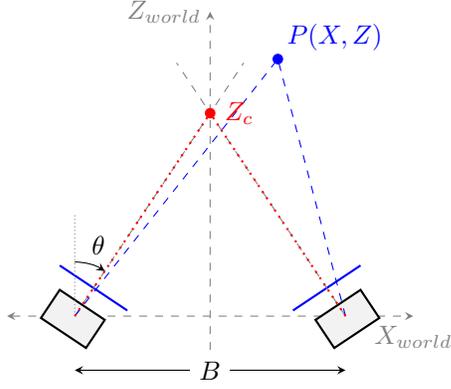
\begin{figure}[t]
    \centering
    \begin{tikzpicture}[scale=0.9, >=stealth]
        \def\B{4.0}          
        \def\Zc{3.0}         
        \def\HalfB{2.0}      
        
        \pgfmathsetmacro{\Theta}{atan(\HalfB / \Zc)}
        
        \coordinate (CL) at (-\HalfB, 0); 
        \coordinate (CR) at (\HalfB, 0);  
        \coordinate (Conv) at (0, \Zc);   
        \coordinate (P) at (1.0, 3.8);    
        
        \draw[->, dashed, gray] (0, -0.5) -- (0, 4.5) node[left] {$Z_{world}$};
        \draw[<->, dashed, gray] (-3, 0) -- (3, 0) node[below] {$X_{world}$};
        
        \begin{scope}[shift={(CL)}, rotate={-\Theta}]
            \draw[thick, fill=gray!10] (-0.4, -0.3) rectangle (0.4, 0.2);
            \draw[thick, blue] (-0.6, 0.5) -- (0.6, 0.5);
            \draw[dashed, gray] (0, 0) -- (0, 4.5);
        \end{scope}
        
        \begin{scope}[shift={(CR)}, rotate={\Theta}]
            \draw[thick, fill=gray!10] (-0.4, -0.3) rectangle (0.4, 0.2);
            \draw[thick, blue] (-0.6, 0.5) -- (0.6, 0.5);
            \draw[dashed, gray] (0, 0) -- (0, 4.5);
        \end{scope}
        
        \draw[dashed, blue] (CL) -- (P);
        \draw[dashed, blue] (CR) -- (P);
        \draw[dotted, thick, red] (CL) -- (Conv); 
        \draw[dotted, thick, red] (CR) -- (Conv);
        
        \filldraw [red] (Conv) circle (2pt) node[right=2pt] {$Z_c$};
        \filldraw [blue] (P) circle (2pt) node[above right] {$P(X, Z)$};
        \draw[<->] (-\HalfB, -0.8) -- (\HalfB, -0.8) node[midway, fill=white] {$B$};
        
        \draw[densely dotted, gray] (CL) -- (-\HalfB, 1.5);
        \draw[->] ($(CL)+(0, 0.8)$) arc (90:{90-\Theta}:0.8);
        \node at ($(CL)+(105-\Theta:1.1)$) {$\theta$};
    \end{tikzpicture}
    \caption{Geometry of a converged stereoscopic system where optical axes intersect at $Z_c$.}
    \label{fig:toedin_geometry}
\end{figure}

Figure~\ref{fig:toedin_geometry} illustrates the converged geometry. The cameras are rotated inward by an angle $\theta$ such that their optical axes intersect at $Z_c$.

\subsubsection{Mathematical Derivation}

We define the world coordinate origin $(0,0,0)$ at the midpoint of the baseline. 

\paragraph{Coordinate Transformation:}
The cameras are translated by $\pm B/2$ and rotated by $\mp \theta$. We first transform the world point $P$ into the local camera coordinate systems $(X_{cam}, Y_{cam}, Z_{cam})$.

\textbf{Left Camera ($L$):}
\begin{equation}
    \begin{bmatrix} X_L \\ Y_L \\ Z_L \end{bmatrix} = 
    \begin{bmatrix}
        \cos\theta & 0 & -\sin\theta \\
        0 & 1 & 0 \\
        \sin\theta & 0 & \cos\theta
    \end{bmatrix}
    \begin{bmatrix} X + B/2 \\ Y \\ Z \end{bmatrix}
\end{equation}

\textbf{Right Camera ($R$):}
\begin{equation}
    \begin{bmatrix} X_R \\ Y_R \\ Z_R \end{bmatrix} = 
    \begin{bmatrix}
        \cos\theta & 0 & \sin\theta \\
        0 & 1 & 0 \\
        -\sin\theta & 0 & \cos\theta
    \end{bmatrix}
    \begin{bmatrix} X - B/2 \\ Y \\ Z \end{bmatrix}
\end{equation}

\paragraph{Projection Equations:}
Applying perspective projection $x = f \cdot X_{cam} / Z_{cam}$ to the transformed coordinates:

\begin{align}
    x_l &= f \frac{(X + B/2)\cos\theta - Z\sin\theta}{(X + B/2)\sin\theta + Z\cos\theta} \\
    x_r &= f \frac{(X - B/2)\cos\theta + Z\sin\theta}{-(X - B/2)\sin\theta + Z\cos\theta}
\end{align}

These equations account for the keystone distortion inherent in toed-in camera configurations.


\end{document}